\documentclass[sigconf,authorversion,nonacm]{acmart}
\newcommand{\etal}{\textit{et al}.}
\newcommand{\eg}{\textit{e}.\textit{g}.}
\usepackage{booktabs} 

\setcopyright{rightsretained}

\acmISBN{978-1-4503-7596-2}

\acmConference[ICVGIP'21]{12th Indian Conference on Computer Vision, Graphics and Image Processing}{December 2021}{Jodhpur, India}
\acmYear{2021}
\copyrightyear{2021}

\acmPrice{15.00}

\editor{Chetan Arora}
\editor{Parag Chaudhuri}
\editor{Subhransu Maji}

\acmDOI{10.1145/3490035.3490275}

\acmArticle{18}

\begin{document}

\title{HDR-cGAN: Single LDR to HDR Image Translation using Conditional GAN}

  \author{Prarabdh Raipurkar}
  \affiliation{%
    \institution{CVIG Lab, IIT Gandhinagar}
    \state{Gujarat}
    \country{India}
  }
  \email{raipurkarprarabdh@gmail.com}
  
  \author{Rohil Pal}
  \affiliation{%
    \institution{CVIG Lab, IIT Gandhinagar}
    \state{Gujarat}
    \country{India}
  }
  \email{rohilpal9763@gmail.com}
  
  \author{Shanmuganathan Raman}
  \affiliation{%
    \institution{CVIG Lab, IIT Gandhinagar}
    \state{Gujarat}
    \country{India}
  }
  \email{shanmuga@iitgn.ac.in}

\renewcommand{\shortauthors}{}

\begin{abstract}
The prime goal of digital imaging techniques is to reproduce the realistic appearance of a scene. Low Dynamic Range (LDR) cameras are incapable of representing the wide dynamic range of the real-world scene. The captured images turn out to be either too dark (underexposed) or too bright (overexposed). Specifically, saturation in overexposed regions makes the task of reconstructing a High Dynamic Range (HDR) image from single LDR image challenging. In this paper, we propose a deep learning based approach to recover details in the saturated areas while reconstructing the HDR image. We formulate this problem as an image-to-image (I2I) translation task. To this end, we present a novel conditional GAN (cGAN) based framework trained in an end-to-end fashion over the HDR-REAL and HDR-SYNTH datasets. Our framework uses an overexposed mask obtained from a pre-trained segmentation model to facilitate the hallucination task of adding details in the saturated regions. We demonstrate the effectiveness of the proposed method by performing an extensive quantitative and qualitative comparison with several state-of-the-art single-image HDR reconstruction techniques.
\end{abstract}

%

\ccsdesc[300]{Computing methodologies~Computational photography}
\ccsdesc[300]{Computing methodologies~Image processing}

\keywords{High Dynamic Range Imaging, Computational Photography, Generative Adversarial Networks}

\maketitle

\section{Introduction}
Traditional camera sensors often struggle to handle real-world lighting contrasts such as bright lamps at night and shadows under the sun. The ability to capture and display both highlights and shadows in the same scene is characterized by the dynamic range of the sensor. Scenes around us offer a vast dynamic range. The Human Visual System (HVS) is very efficient in capturing a good portion of this dynamic range, due to which we can see a large variation in intensity (brightness). Traditional cameras have a limited dynamic range and are thus known as Low Dynamic Range (LDR) cameras. Images captured by these cameras often suffer from information loss in over and under exposed regions.
The recent advancements in High Dynamic Range (HDR) imaging have shown to overcome these issues and match the dynamic range of the HVS. There are specialized hardware devices \cite{855857,1467255} that can create HDR content directly, but they are very expensive and inaccessible to the common user. The most common strategy for creating HDR content is to capture a sequence of LDR images at three different exposures and then merge them \cite{10.1145/258734.258884,Mann95onbeing,article2,YAN2017160}. These methods produce fine HDR images in static scenes but are prone to ghosting and blurring artifacts when the scene is dynamic. Due to these reasons, single image dynamic range expansion methods have gained popularity recently. This task is challenging because HDR images are typically stored in 32-bit floating point format, while input LDR intensity values are usually 8-bit unsigned integers.

Due to their data-driven nature, deep learning-based HDR reconstruction approaches \cite{eilertsen2017hdr,10.1145/3130800.3130834,liu2020singleimage,marnerides2018expandnet,Santos2020SingleIH} have gained popularity lately. These techniques use one or more encoder-decoder architectures to model various components of the LDR-HDR pipeline. Marnerides \etal \cite{10.1145/3130800.3130834} follow an indirect approach by generating a multi-exposure stack from a single LDR image and then merging it using \cite{10.1145/258734.258884}. Eilertsen  \etal \cite{eilertsen2017hdr} use a single U-Net \cite{inproceedings} like architecture to hallucinate the missing details in saturated areas. Santos \etal \cite{Santos2020SingleIH} improve upon \cite{eilertsen2017hdr} by introducing a feature masking mechanism. Marnerides \etal \cite{marnerides2018expandnet} employ multiple fully convolutional architectures to focus on different levels of details in the LDR image while Liu \etal \cite{liu2020singleimage} train separate networks to model the sub-components of the LDR-HDR pipeline.

We treat the single-image based HDR reconstruction problem as an image-to-image (I2I) translation task which learns the mapping between the LDR and HDR domains. Generative Adversarial Networks (GANs) \cite{goodfellow2014generative} are commonly used generative models in I2I tasks. Therefore, we leverage a conditional GAN (cGAN) based framework \cite{pix2pix2017,mirza2014conditional}. It is very challenging to model the LDR-HDR translation using a single generic architecture \cite{liu2020singleimage}. Hence, our generator pipeline comprises three modules: (1) \textbf{Linearization module} converts the non-linear input LDR image to linear irradiance output, (2) \textbf{Over-exposed region correction (OERC) module} focuses on hallucinating saturated pixel values by estimating an over-exposure mask using \cite{inbook}, and (3) \textbf{Refinement module} aids in rectifying the irregularities in the output of the OERC module. The Linearization module recovers the missing details in the underexposed and normally-exposed regions of the image whereas the OERC module predicts the missing details in the overexposed regions. For all the three modules, the encoder-decoder architecture is inspired by Attention Recurrent Residual Convolutional Neural Network \cite{alom2018recurrent,oktay2018attention}. The framework is trained end-to-end using a combination of adversarial, reconstruction and perceptual losses. Reconstruction loss is a pixel-wise L1 norm that captures the low-level details in the HDR image. Perceptual loss \cite{johnson2016perceptual} is an L2 norm between the VGG-16 \cite{simonyan2015deep} deep feature vectors of the ground truth HDR and the reconstructed HDR image. cGAN based adversarial loss automatically captures high-level details through a min-max game between generator and discriminator. The contributions of our work are outlined as follows:
\begin{itemize}
    \item We propose a novel conditional GAN based framework to reconstruct an HDR image from a single LDR image.
    \item We formulate the LDR-HDR pipeline as a sequence of three steps, linearization, over-exposure correction, and refinement.
    \item We leverage a threshold-free mask mechanism to recover details in the saturated regions of the input LDR image.
    \item We use a combination of adversarial, reconstruction and perceptual losses to reconstruct sharp details in the output HDR image.
    \item Our quantitative and quantitative assessments demonstrate that our method performs comparably to the state-of-the-art approaches.
\end{itemize}

We present the rest of the paper as follows. Section \ref{related_work} outlines the related work. In Section \ref{ch:method1}, we describe our methodology and the proposed framework in detail. This is followed by dataset and implementation details in Section \ref{training}. We showcase our experiments, results and limitations of our method in Section \ref{experiments} and finally conclude the paper in Section \ref{conclusion}.

\section{Related work}
\label{related_work}
\subsection{Multi-image HDR reconstruction}
The typical method of capturing the entire dynamic range of a scene is merging a stack of bracketed exposure LDR images \cite{10.1145/258734.258884,Mann95onbeing}. Multi-stack approaches heavily rely on proper alignment of the neighboring LDR images with the middle one in order to avoid artifacts in the HDR output. Recent methods apply CNNs to model the alignment and fusion of the multi-exposure stack. Kalantari \etal \cite{LearningHDR} proposed the first deep learning approach to reconstruct an HDR image from a multi-exposure stack of a dynamic scene. It relies on optical flow estimation to align the input images. Wu \etal \cite{10.1007/978-3-030-01216-8_8} introduced a framework composed of multiple encoders and a single decoder to fuse the input images without explicit alignment using optical flow. Yan \etal \cite{Yan2019AttentionGuidedNF} improved upon \cite{10.1007/978-3-030-01216-8_8} by using an attention network to align the input LDR images. In \cite{10.1145/2366145.2366222}, Sen \etal formulated this problem as an energy-minimization problem and introduced a novel patch-based algorithm that optimizes the alignment and reconstruction process jointly. \cite{Niu2021HDRGANHI} is a recent method that uses a GAN-based model to fuse the bracketed exposure images into an HDR image.

\subsection{Single-image HDR reconstruction}
This problem is also known as inverse tone-mapping (iTMO) \cite{DBLP:conf/graphite/BanterleLDC06}. Although this class of approaches does not suffer from ghosting and blurring artifacts like the multi-exposure techniques, this is a challenging problem as one has to infer the lost details in both the over and under exposed regions from a single arbitrarily exposed LDR image. 

The first class of techniques generates multiple LDR exposures from a single LDR image and then fuses them using popular merging techniques \cite{10.1145/258734.258884,10.1109/PG.2007.23}. Endo \etal \cite{10.1145/3130800.3130834} and Lee \etal \cite{8457442} generated bracketed exposure images from a single LDR image using fully-convolutional networks and then merged these images using Debevec and Malik's algorithm \cite{10.1145/258734.258884} to create the final HDR image. Lee \etal \cite{lee2018deep} used a conditional GAN model to generate the multi-exposure stack from a single LDR image and then fused it using \cite{10.1145/258734.258884}.

The second class of techniques learns a direct mapping between the input LDR image and ground truth HDR image. Eilertsen \etal \cite{eilertsen2017hdr} reconstructed saturated areas in the LDR image using an encoder-decoder network pre-trained on a HDR dataset sampled from the MIT places dataset \cite{NIPS2014_3fe94a00}. Marnerides \etal \cite{marnerides2018expandnet} introduced a 3-branch CNN architecture comprising local, dilation, and global branches responsible for capturing local-level, medium-level, and high-level details, respectively. The outputs of these branches are finally fused using convolution layers to produce the HDR image. Santos \etal \cite{Santos2020SingleIH} employed a feature masking mechanism to mitigate the contribution of saturated areas while computing the HDR output. In order to create realistic textures in the output, the authors pre-trained the network on an inpainting task and used VGG-based perceptual loss in the finetuning stage. Yang \etal \cite{Yang_2018_CVPR} reconstructed HDR image and then estimated a tonemapped version using a sequence of CNNs. Liu \etal \cite{liu2020singleimage} explicitly modelled the LDR-HDR pipeline using multiple sub-networks and then trained them in a stage-wise manner. Khan \etal \cite{8969167} reconstructed the HDR output in an iterative manner using a feedback block. In their proposed network, Li and Fang \cite{li2019hdrnet} apply channel-wise attention to model the feature dependencies between the channels. Some recent approaches such as \cite{moriwaki2018hybrid} employ adversarial training for reconstructing HDR images.

\section{Single LDR to HDR Translation}
\label{ch:method1}
This section describes the overall flow of our approach to reconstruct an HDR image from a single LDR image. In Section \ref{promethod}, we give a broad overview of the proposed method. Then, we describe in detail the generator pipeline in Section \ref{genny}. In Section \ref{sec:loss}, we explain the loss functions used to train the model.

\subsection{Methodology}
\label{promethod}

\begin{figure}[h]
\begin{center}
  \includegraphics[width=0.9\linewidth]{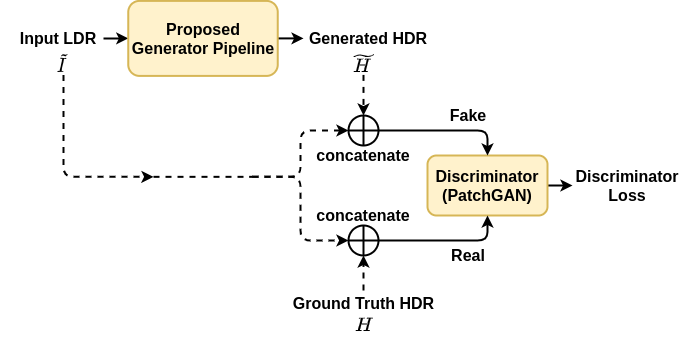}
\end{center}
  \caption{Overview of our cGAN based approach.}
\label{fig:long}
\label{fig:pix2pix}
\end{figure}

The goal of our work is to translate an input LDR image into a high-quality HDR image. We formulate our single LDR to HDR conversion problem as an image-to-image translation problem. The image-to-image translation task is the learning of mapping between two different image domains \cite{pang2021imagetoimage}. CNN-based approaches attempt to minimize the hand-engineered loss functions to learn these mappings. Designing a specific loss function for the single LDR to HDR conversion task is still an open problem. Generative Adversarial Networks (GANs) \cite{goodfellow2014generative} are very well-known for automatically learning a loss function to make output indistinguishable from the target distribution. So, we employ a conditional GAN (cGAN) \cite{mirza2014conditional} based approach where the generation of the output image is conditional on the source image. Unlike unconditional GANs, the discriminator also observes the input image in cGANs. We utilize pix2pix architecture \cite{pix2pix2017} which is a strong baseline built on cGAN for many I2I tasks. We propose a generator pipeline and a combination of losses which consists of adversarial, reconstruction and perceptual losses. The outline of our method is shown in Figure \ref{fig:pix2pix}.

Input LDR image $\widetilde{I}$ is fed to the proposed generator pipeline to output an HDR image $\widetilde{H}$. Then, $\widetilde{H}$ is concatenated with $\widetilde{I}$ and fed to the discriminator with a fake label to get fake prediction loss. Similarly, the ground truth HDR $H$ is concatenated with $\widetilde{I}$ and fed to the discriminator but with a real label to get real prediction loss. The total discriminator loss is the average of fake prediction loss and real prediction loss. We use a convolutional ``PatchGAN" classifier \cite{pix2pix2017} as our discriminator network. PatchGAN works by classifying each $N \times N$  patch in the image as “real vs. fake” and then takes an average to classify the whole image as real or fake. We keep 70 $\times$ 70 as patch size in our implementation.

\subsection{Proposed Generator pipeline}
\label{genny}
\begin{figure*}[h]
\begin{center}
\includegraphics[width=0.9\linewidth]{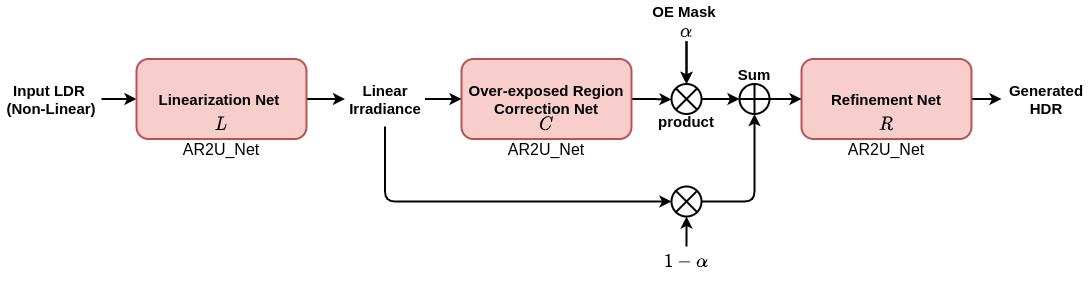}
\end{center}
  \caption{Our Proposed Generator Pipeline.}
\label{fig:xx}
\end{figure*}

Figure \ref{fig:xx} shows the proposed generator pipeline. The whole generator pipeline is trained in an end-to-end fashion. The pipeline consists of three subnetworks: Linearization Net $(L)$, Over-exposed Region Correction Net $(C)$ and Refinement Net $(R)$. Although the purpose served by each subnetwork is different, the architecture involved in all the three subnetworks is the same [$AR2U\_Net$]. The subnetwork architecture is discussed later in Section \ref{subsec:Subnetwork Architecture}. We will now discuss the purpose served by individual network modules.

\textbf{Linearization Module.} Camera imaging systems apply a non-linear function also known as camera response function (CRF) to shrink the broad range of scene irradiance values into a fixed range of intensity values \cite{1323796}. The goal of single-image based HDR reconstruction is to learn the inverse mapping from non-linear image intensity to actual scene irradiance. This non-linear mapping is also known as inverse camera response function (inverse CRF). Since camera companies do not provide CRF and consider it their proprietary information, there is no prior information about the actual CRF in our problem setting. The Linearization Net $L$ tries to roughly estimate the inverse CRF and convert non-linear LDR input $\widetilde{I}$ into linear image irradiance $\widetilde{E}$ as depicted in Equation \eqref{lnet}. This module is able to recover details in the normally-exposed and the underexposed regions to a good extent.
\begin{equation}
    \widetilde{E} = L(\widetilde{I})
    \label{lnet}
\end{equation}

\textbf{Over-exposed Region Correction (OERC) Module.} Linearization Module is able to transfer the LDR input to the required linear domain, but it still struggles to recover the image information in overexposed regions. Missing details in these regions immensely reduce the visual appeal of an image. The role of the OERC Module is to predict these details. The corrected image $\widetilde{M}$ is calculated using Equation \eqref{cnet} where $\widetilde{E}$ is the image irradiance, $C$ is the Over-exposed Region Correction Net, $\alpha$ is the Over Exposure mask (OE Mask) and $\odot$ denotes element-wise multiplication. The corrected image $\widetilde{M}$ is obtained by combining the content of network output in the overexposed regions and the image irradiance in the well-exposed regions. 
\begin{equation}
    \widetilde{M} = \alpha \odot C(\widetilde{E}) + (1-\alpha) \odot \widetilde{E}
    \label{cnet}
\end{equation}

\begin{figure}[h]
\begin{center}
   \includegraphics[width=0.9\linewidth]{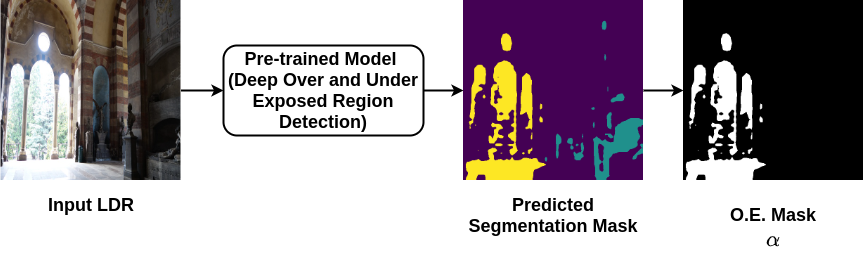}
\end{center}
   \caption{Extracting Over Exposure Mask from LDR input image.}
\label{fig:oemask}
\end{figure}

The OE Mask $\alpha$ is the key component of the OERC module. Traditional methods use an intensity threshold to detect the overexposed regions in an image. The major drawback of using thresholds is that there is no clear distinction between actual whites and washed-out regions. Jain and Raman \cite{inbook} proposed a threshold-free model for over and under exposed region detection. \cite{inbook} utilized a pre-trained DeepLabv3 \cite{chen2017rethinking} model for image segmentation. Starting with the pre-trained weights, they retrained the model with their hand-curated dataset to classify each pixel into three segmentation classes, viz.: over, under and normally exposed pixels. We directly perform the inference on this model and obtain the OE Mask $\alpha$ by representing overexposed pixels with label $1$ and both under and normally exposed pixels with label $0$. Figure \ref{fig:oemask} shows the OE Mask obtained by inferring the model on an input LDR image. The resulting OE Mask $\alpha$ indeed aids our approach of recovering missing contents in overexposed regions.

\textbf{Refinement Module.} Modern camera pipelines are more complex than just applying CRFs to obtain non-linear digital intensities. High-end cameras employ operations such as demosaicing, white
balancing, noise reduction and gamut mapping for high-quality image
formation \cite{InCamera,karaimer_brown_ECCV_2016,548d4cc}. Digital cameras suffer from transmission errors of the imaging system's optics and fixed pattern noises \cite{1323796}. These errors are assumed to be linear with respect to scene irradiance. So, we deal with these errors with the Refinement Net $R$ as per Equation \eqref{rnet}, 
\begin{equation}
    \widetilde{H} = R(\widetilde{M})
    \label{rnet}
\end{equation}

where $\widetilde{M}$ is the corrected image obtained from OERC Module and $\widetilde{H}$ is the final generator output (linear HDR image). We observe visible color artifacts in under and normally exposed regions if we omit the Refinement module from our generator pipeline.

In summary, the generator transfers the input 8-bit LDR image $\widetilde{I}$ to the output 32-bit floating point HDR image $\widetilde{H}$. The workflow of the whole generator pipeline is depicted in Equation \ref{overallnet} where $\Phi$ denotes the generator pipeline consisting of Linearization Net $L$, Over-exposed Region Correction Net $C$ and Refinement Net $R$.\;$\alpha$ denotes the Over Exposure mask and $\odot$ denotes element-wise multiplication. 
\begin{equation}
    \widetilde{H} = \Phi(\widetilde{I}) = R\;	(\alpha \odot C(L(\widetilde{I})) \;	+\;	 (1-\alpha) \odot L(\widetilde{I}))
    \label{overallnet}
\end{equation}

\subsubsection{Subnetwork Architecture}
\label{subsec:Subnetwork Architecture}

\begin{figure*}[h]
\begin{center}
   \includegraphics[width=0.9\linewidth,height=7cm]{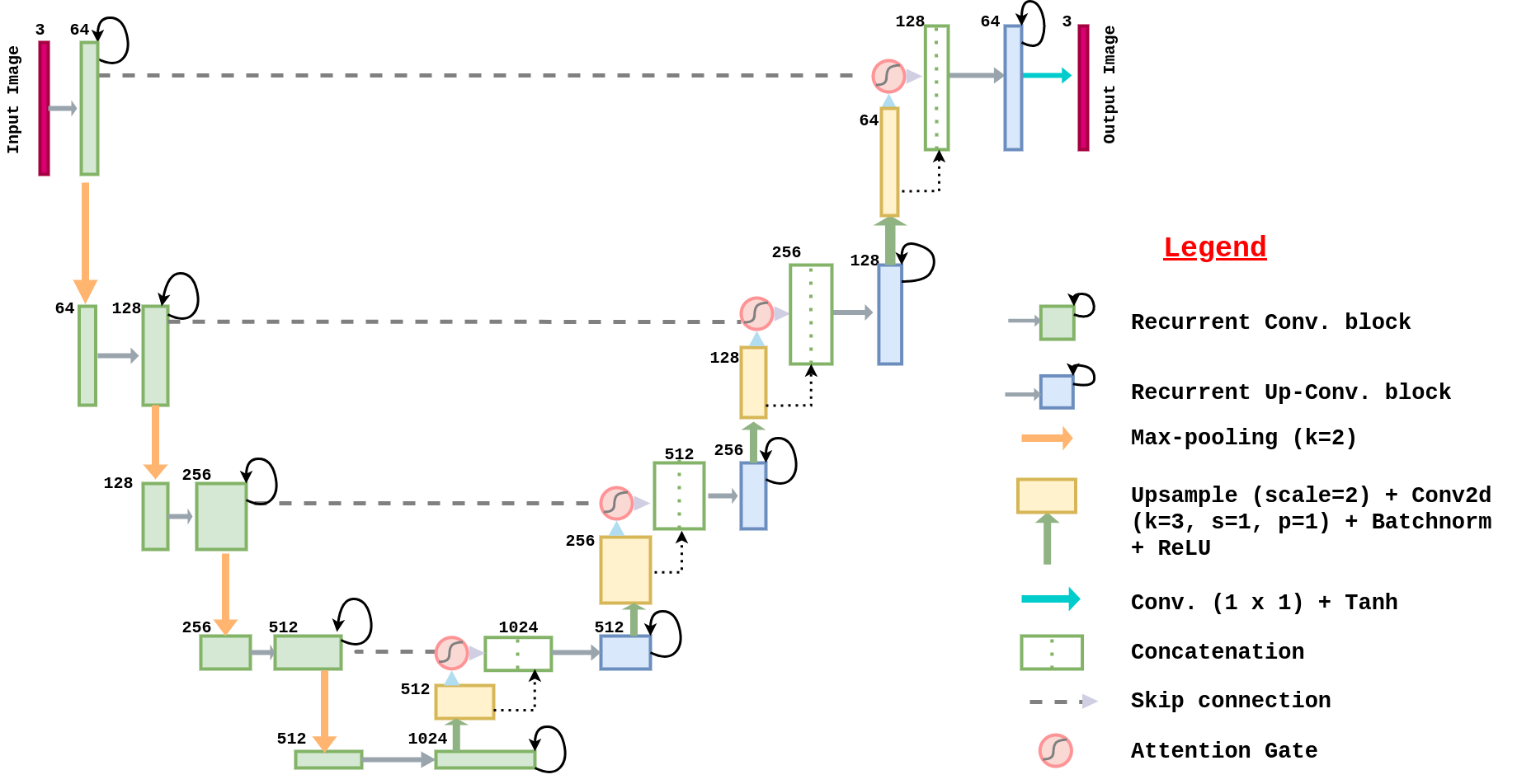}
\end{center}
   \caption{Attention Recurrent Residual UNet architecture [$AR2U\_Net$] \cite{imagesegmentation}.}
\label{fig:long}
\label{fig:AttR2UNet}
\end{figure*}

Figure \ref{fig:AttR2UNet} shows the architecture of Attention Recurrent Residual Convolutional Neural Network based on U-Net, commonly known as Attention R2U-Net [$AR2U\_Net$]. Each of the subnetworks in our generator pipeline (i.e., Linearization Net, Over-exposed Region Correction Net and Refinement Net) adopt the architecture of Attention R2U-Net. Attention R2U-Net is the integration of two advanced works in deep neural network architectures: Recurrent Residual U-Net \cite{alom2018recurrent} and Attention U-Net \cite{oktay2018attention}. The general structure is almost similar to U-Net with encoder, decoder and long skip connections.

As shown in Figure \ref{fig:AttR2UNet}, the input is first passed through Recurrent Residual block layers with $64$ filters in the last convolutional layer, and then progressively downsampled until a bottleneck layer with $1024$ filters. Recurrent block accumulates the features with respect to different time-steps and thus learns better features for the model without increasing the number of network parameters \cite{7298958}. Attention block learns a scaling factor for input features propagated through skip connections. It essentially learns to highlight the salient features specific to the task, and then the learned features are concatenated to the expansive path where the feature maps are progressively upsampled. These recurrent, residual and attention blocks indeed have a compelling effect on the performance of the model.

\subsection{Loss function}
\label{sec:loss}
HDR images represent the scene radiance in the 32-bit floating point format. To simulate the scene appearance of nearly the same quality on the 8-bit encoded normal displays, HDR images need to be tonemapped. Tonemapping boosts the intensity level in dark pixels and suppresses the bright pixels. Defining a loss function directly in the linear HDR domain may not compensate for the errors in the dark regions. Thus, we use the $\mu$ law \cite{LearningHDR} as a tonemapping operator which compresses the range of HDR image $H$ as per the Equation \ref{mulaw} where $\mu$ defines the amount of compression and $\mathcal{T}(H)$ denotes the tonemapped image. We set $\mu=5000$ in our implementation.
\begin{equation}
    \mathcal{T}(H) = \frac{log(1+\mu H)}{log(1+\mu)}
    \label{mulaw}
\end{equation}

We calculate all the loss functions between the tonemapped generated HDR output and the tonemapped ground truth HDR. Specifically, we use a combination of three loss functions for training the model. Equation \ref{weightedloss} depicts our final loss function, which is a weighted summation of adversarial loss, reconstruction loss and perceptual loss. We use $\lambda_1 = 100$ and $\lambda_2 = 0.005$ in our implementation.
\begin{equation}
    \mathcal{L} =  \mathcal{L}_{cGAN} + \lambda_1(\mathcal{L}_{Rec} + \lambda_2 \mathcal{L}_P)
    \label{weightedloss}
\end{equation}

\textbf{Conditional Adversarial loss ($\mathcal{L}_{cGAN}$).} Our cGAN model is conditioned by feeding an input LDR image $x$ in both the generator $G$ and discriminator $D$. Input $x$ and prior noise $z$ are combined in a joint hidden representation and given as input to the generator. In the discriminator, the   tonemapped ground truth HDR $\mathcal{T}(H)$ and tonemapped generated HDR $\mathcal{T}(\widetilde{H})$ [$=\mathcal{T}(G(x,z))$] are fed along with $x$ as input to obtain total discriminator loss as shown in Figure \ref{fig:pix2pix}. Equation \ref{adv1loss} \cite{pix2pix2017} shows the objective function of our conditional GAN.
\begin{equation}
\begin{aligned}
\mathcal{L}_{cGAN}(G,D) =& \mathbb{E}_{x,\mathcal{T}(H)}[log D(x,\mathcal{T}(H))] \\
& + \mathbb{E}_{x,z} [log(1-D(x,\mathcal{T}(G(x,z)))]
\label{adv1loss}
\end{aligned}
\end{equation}

Generator $G$ tries to minimize this objective function, whereas discriminator $D$ tries to maximize it against the generator. This min-max game can be depicted as per the Equation \ref{advloss}.
\begin{equation}
    G^\ast = \arg\min_{G}\max_{D} \;\mathcal{L}_{cGAN}(G,D) 
    \label{advloss}
\end{equation}

\textbf{Reconstruction loss ($\mathcal{L}_{Rec}$).} Since the aim of generator training is to produce the output HDR image close to the ground truth HDR image; we employ pixel-wise $L1$ distance loss as our HDR reconstruction loss. $L1$ norm produces less blurry results as compared to $L2$ norm and is robust to the outlier pixel values. The linear HDR luminance values may range from very low (dark regions) to very high (very bright regions). Tonemapping operators smoothly boost the dark image regions. So, we define the HDR reconstruction loss between tonemapped generated HDR image and tonemapped ground truth HDR image as follows
\begin{equation}
    \mathcal{L}_{Rec} = {\| \mathcal{T}(\widetilde{H}) - \mathcal{T}(H) \|} _1
    \label{recloss}
\end{equation}

where $\widetilde{H}$ denotes the generated HDR output, $H$ denotes the ground truth HDR image and $\mathcal{T}(.)$ is the $\mu$ law tonemapping operator.

\textbf{Perceptual loss ($\mathcal{L}_{P}$).} To generate more plausible and realistic details in an output HDR image, we apply perceptual loss \cite{johnson2016perceptual}. This loss attempts to be closer to perceptual similarity for human eyes by measuring the high-level semantic differences. The perceptual loss function can be expressed as
\begin{equation}
    \mathcal{L}_{P} = \sum_{l} {\| \; \phi_l(\mathcal{T}(\widetilde{H})) - \phi_l(\mathcal{T}(H)) \; \|} _2
    \label{perloss}
\end{equation}

where $\phi_l(.)$ denotes the feature map output of $l^{th}$ pooling layer in VGG-16 \cite{simonyan2015deep} network. Here, the $L2$ distance loss is calculated between the feature maps of the tonemapped generated HDR image and the tonemapped ground truth HDR image.

\begin{table*}[hbt!]
\begin{center}
\begin{tabular}{|c||l|l|l||l|l|l|}
\hline
\multicolumn{1}{|l||}{} & \multicolumn{3}{c||}{\textbf{HDR-EYE}}                                                                     & \multicolumn{3}{c|}{\textbf{HDR-REAL}}                                                                    \\ \hline
Method                 & \multicolumn{1}{c|}{PSNR$(\uparrow)$} & \multicolumn{1}{c|}{SSIM$(\uparrow)$} & \multicolumn{1}{c||}{VDP $2.2$$(\uparrow)$} & \multicolumn{1}{c|}{PSNR$(\uparrow)$} & \multicolumn{1}{c|}{SSIM$(\uparrow)$} & \multicolumn{1}{c|}{VDP $2.2$$(\uparrow)$} \\ \hline\hline
DrTMO \cite{10.1145/3130800.3130834} & \;$9.28\pm2.98$ & $0.69\pm0.15$ & $48.33\pm5.16$ & \;$5.54\pm3.55$ & $0.36\pm0.26$ & $43.64\pm6.51$ \\ \hline
HDRCNN \cite{eilertsen2017hdr} & $16.12\pm3.77$ & $0.74\pm0.12$ & $50.75\pm5.75$ & $13.34\pm7.68$ & $0.53\pm0.29$ & $47.20\pm7.77$ \\ \hline
ExpandNet \cite{marnerides2018expandnet} & \color{blue}{$17.12\pm4.27$} & \color{red}{$0.79\pm0.13$} & $51.09\pm5.87$ & $12.84\pm6.90$ & $0.50\pm0.29$ & $46.70\pm7.90$\\ \hline
SingleHDR \cite{liu2020singleimage} & $15.47\pm6.65$ & $0.71\pm0.19$ &  \color{red}{$53.15\pm5.91$} & \color{red}{$19.07\pm7.16$} & \color{red}{$0.64\pm0.28$} &  \color{red}{$50.13\pm7.74$}\\ \hline
Ours & \color{red}{$17.57\pm4.68$} & \color{blue}{$0.78\pm0.15$} & \color{blue}{$51.94\pm5.56$} & \color{blue}{$16.06\pm5.87$} & \color{blue}{$0.57\pm0.31$} & \color{blue}{$48.35\pm6.73$} \\ \hline
\end{tabular}
\end{center}
\caption{Quantitative evaluation of our approach. $(\uparrow)$ indicates higher the value of metric better is the reconstruction quality. {\color{red}{Red}} indicates the best and {\color{blue}{blue}} indicates the second-best performing method for each metric.}
\label{hdreyerealtable}
\end{table*}

\section{Training}
\label{training}
\subsection{Dataset}
Training of our cGAN model requires a dataset having a large number of LDR-HDR image pairs. To make our model robust to different CRFs, we train our model on the publicly available HDR-REAL and HDR-SYNTH datasets \cite{liu2020singleimage}. The HDR-REAL training set consists of 9786 LDR-HDR pairs captured using 42 different camera models. Additionally, we use the HDR-SYNTH dataset, which is a collection of 496 HDR images used in previous works \cite{article1, Funt2010TheRO, 10.1117/12.845394, article2, Ward06highdynamic, inproceedings1} and online sources such as Pfstools HDR gallery\footnote{\url{http://pfstools.sourceforge.net/hdr_gallery.html}}. As \cite{liu2020singleimage} does not provide the input LDR images in HDR-SYNTH dataset, we synthetically generate LDR samples from these HDR images using Equation \eqref{l2hclip} \cite{https://doi.org/10.1111/cgf.13630}, where $\widetilde{H}$ is the HDR image in the linear domain, $t$ is the exposure time and $Z$ is the synthetically generated LDR sample. 
\begin{equation}
    Z = f(\widetilde{H}) = \mathrm{clip}[(\widetilde{H} t)^{1/\gamma}]  
    \label{l2hclip}
\end{equation}

We use a gamma curve with $\gamma$ = 2.2 as an approximation of the CRF, which transforms the linear HDR image $\widetilde{H}$ into a non-linear LDR domain $Z$. Clip function limits the LDR output in range [0,1]. We select 4 exposure values  $(t = 2^{0.5}, 2^1, 2^2, 2^4)$ to generate 4 LDR images for each HDR image. These values provide a good exposure variation from underexposed to highly overexposed LDR images. Combining HDR-REAL and HDR-SYNTH, our training dataset in total consists of 11770 $(= 9786 + (496 \times 4) )$ LDR-HDR image pairs.

\subsection{Implementation Details} The proposed method is implemented under the PyTorch \cite{paszke2019pytorch} framework. We follow the conditional GAN (cGAN) training framework as suggested in \cite{pix2pix2017}, where a generator and a discriminator are updated alternately. All weights of the network are initialized using the normal distribution. We train the whole network with an initial learning rate of $2\times10^{-4}$ for the first $100$ epochs and then adopt a linear decay policy for another $100$ epochs. For loss optimization, we use Adam optimizer \cite{Kingma2015AdamAM} with hyper-parameters $\beta_{1}=0.5$ and $\beta_{2}=0.999$. The original $512\times512$ images are resized to $256\times256$ for the training purpose. The general training with a batch size of $1$ takes about six days on a PC with an NVIDIA GeForce RTX 2080 Ti GPU. 

\begin{figure*}[h]
\;\;Input LDR \;\;\;\;\;\;\;\;\;\;  HDRCNN\cite{eilertsen2017hdr} \;\;\;\;\;\;\;\;\; ExpandNet\cite{marnerides2018expandnet} \;\;\;\;\;\;\;\; SingleHDR\cite{liu2020singleimage}\;\;\;\;\;\;\;\;\;\;\;\;\;\;\;\;  Ours \;\;\;\;\;\;\;\;\;\;\;\;\;\;\;  Ground truth\\
\vspace{-0.5mm}
\begin{center}
    \makebox[0pt][r]{\makebox[30pt]{\raisebox{30pt}{\rotatebox[origin=c]{0}{(a)}}}}%
    \includegraphics[width=0.15\linewidth, height=2.5cm]{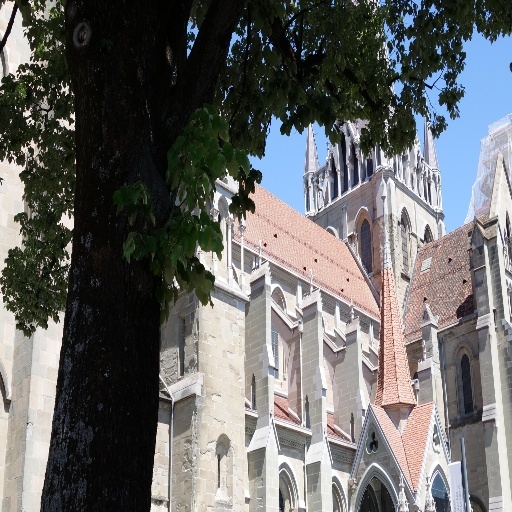}
    \includegraphics[width=0.15\linewidth, height=2.5cm]{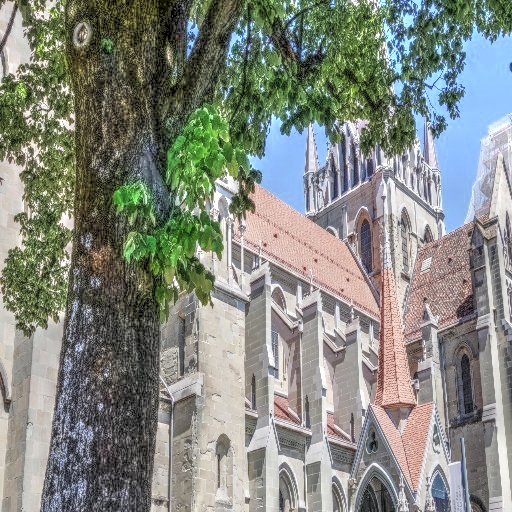}
    \includegraphics[width=0.15\linewidth, height=2.5cm]{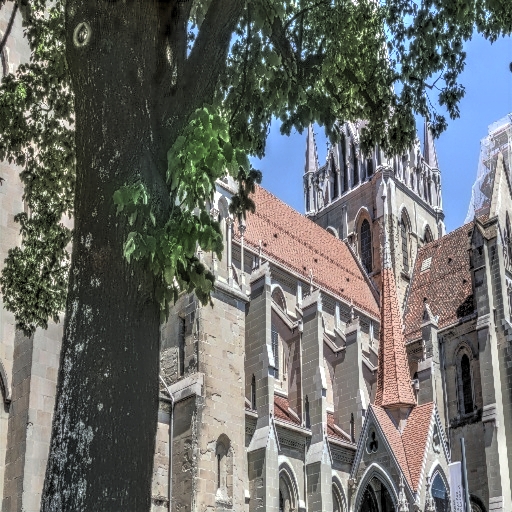}
    \includegraphics[width=0.15\linewidth, height=2.5cm]{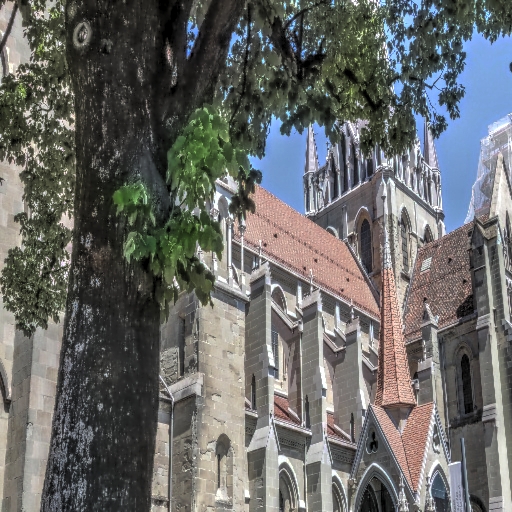}
    \includegraphics[width=0.15\linewidth, height=2.5cm]{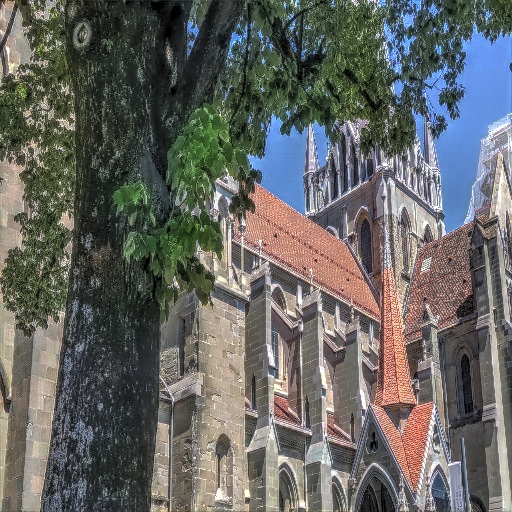}
    \includegraphics[width=0.15\linewidth, height=2.5cm]{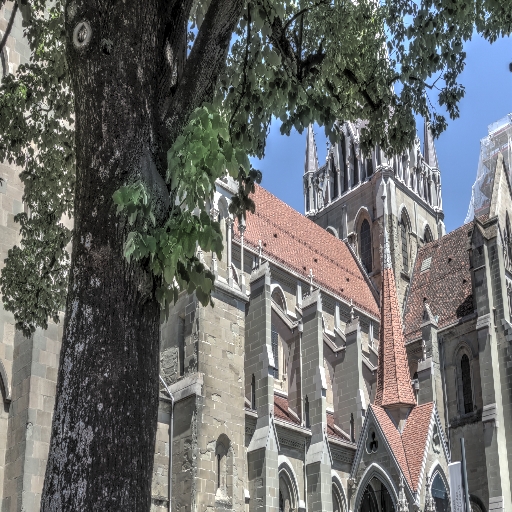}\\
    
    \vspace{0.5mm}
    \makebox[0pt][r]{\makebox[30pt]{\raisebox{30pt}{\rotatebox[origin=c]{0}{(b)}}}}%
    \includegraphics[width=0.15\linewidth, height=2.5cm]{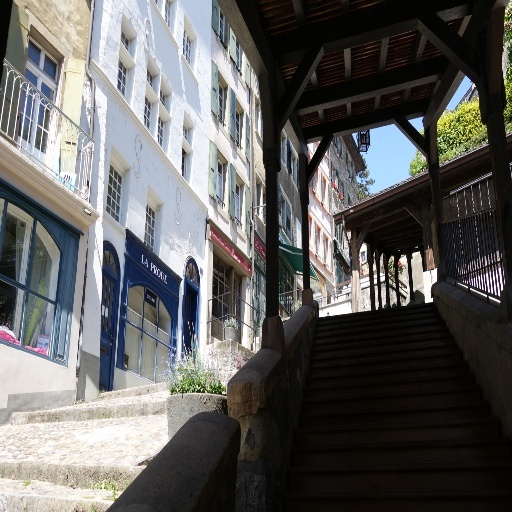}
    \includegraphics[width=0.15\linewidth, height=2.5cm]{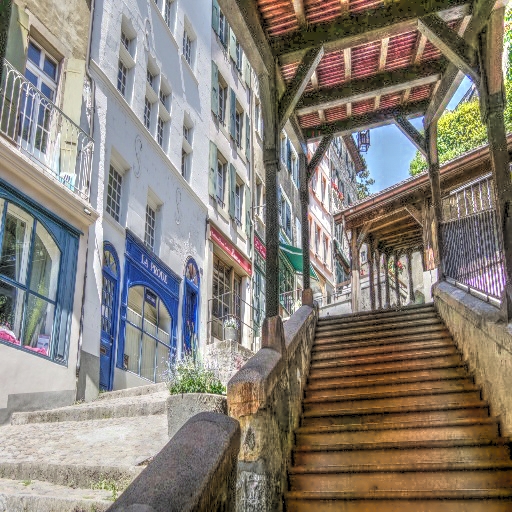}
    \includegraphics[width=0.15\linewidth, height=2.5cm]{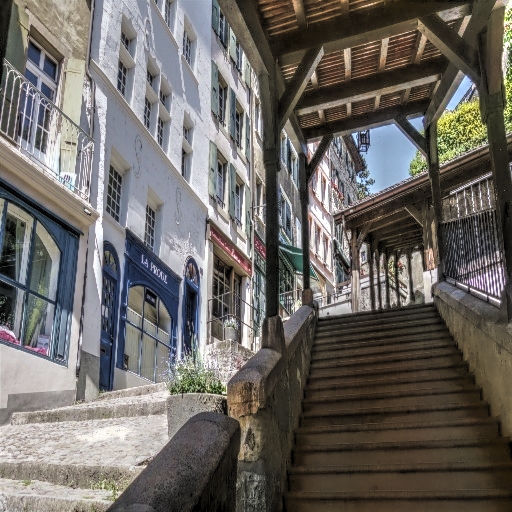}
    \includegraphics[width=0.15\linewidth, height=2.5cm]{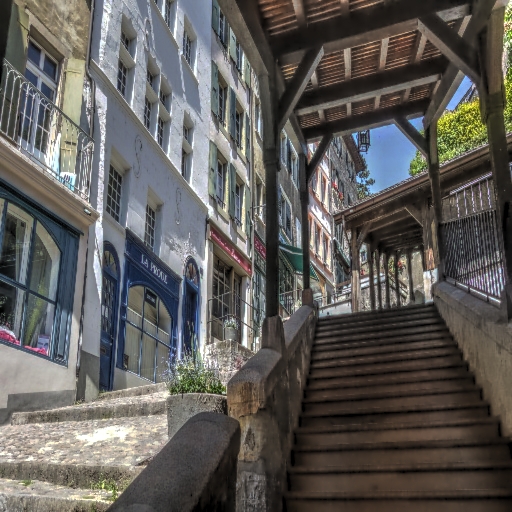}
    \includegraphics[width=0.15\linewidth, height=2.5cm]{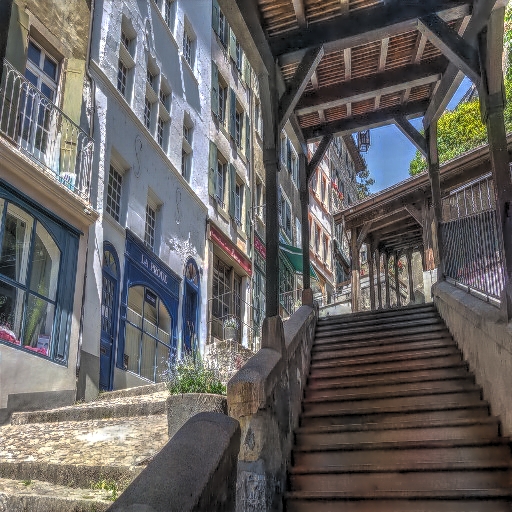}
    \includegraphics[width=0.15\linewidth, height=2.5cm]{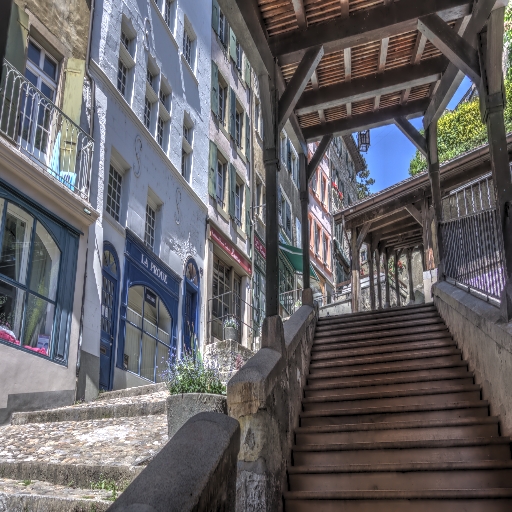}\\
    
    \vspace{0.5mm}

    \makebox[0pt][r]{\makebox[30pt]{\raisebox{30pt}{\rotatebox[origin=c]{0}{(c)}}}}%
    \includegraphics[width=0.15\linewidth, height=2.5cm]{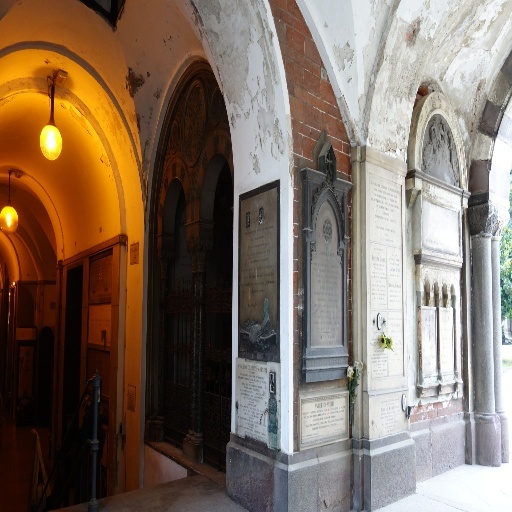}
    \includegraphics[width=0.15\linewidth, height=2.5cm]{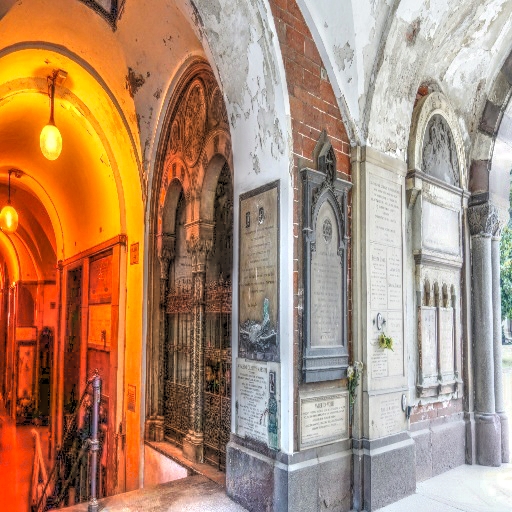}
    \includegraphics[width=0.15\linewidth, height=2.5cm]{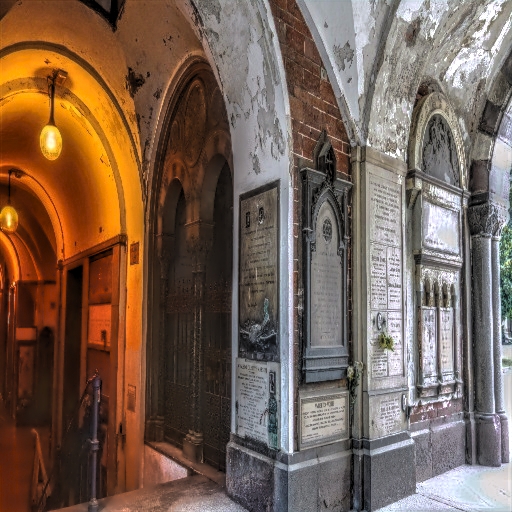}
    \includegraphics[width=0.15\linewidth, height=2.5cm]{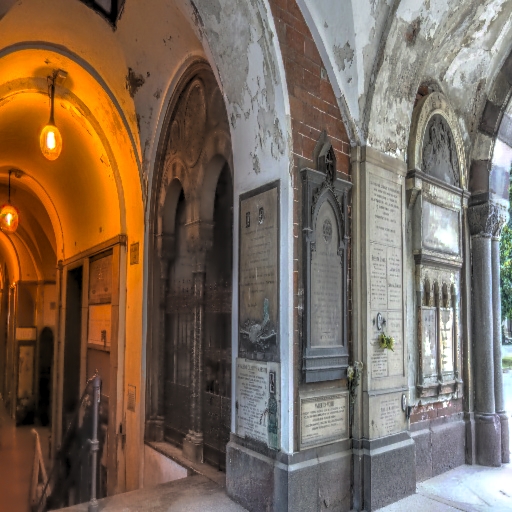}
    \includegraphics[width=0.15\linewidth, height=2.5cm]{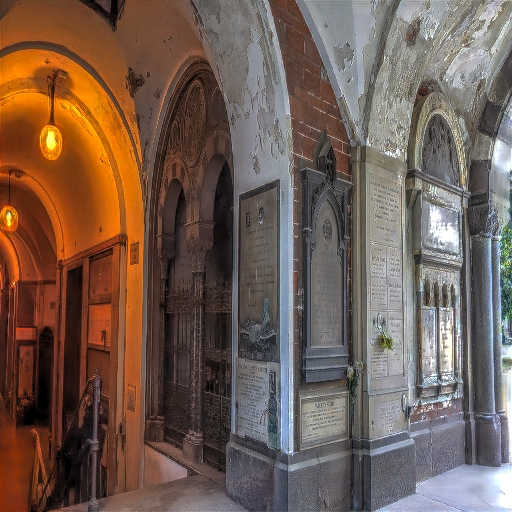}
    \includegraphics[width=0.15\linewidth, height=2.5cm]{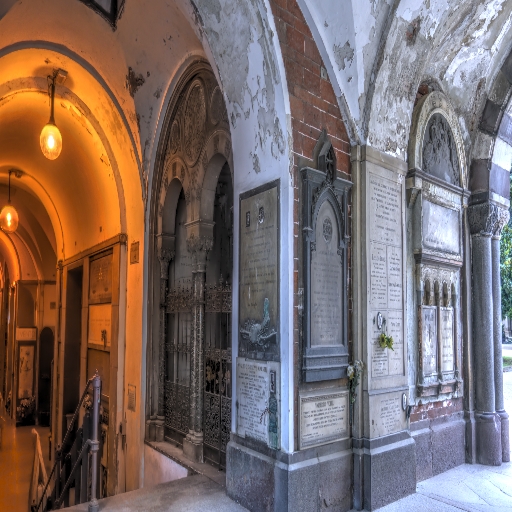}\\
    
    \vspace{0.5mm}
    
    
    
    
    
    \makebox[0pt][r]{\makebox[30pt]{\raisebox{30pt}{\rotatebox[origin=c]{0}{(d)}}}}%
    \includegraphics[width=0.15\linewidth, height=2.5cm]{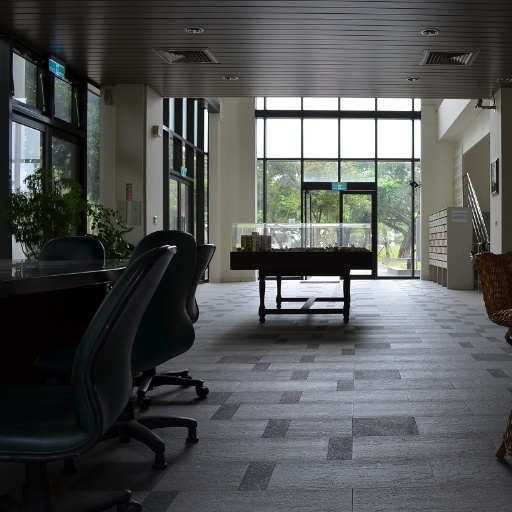}
    \includegraphics[width=0.15\linewidth, height=2.5cm]{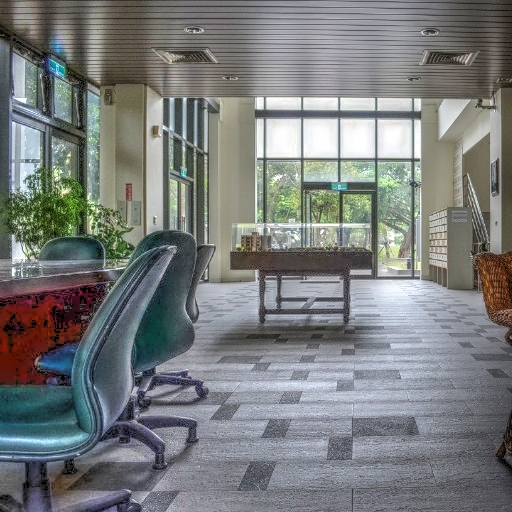}
    \includegraphics[width=0.15\linewidth, height=2.5cm]{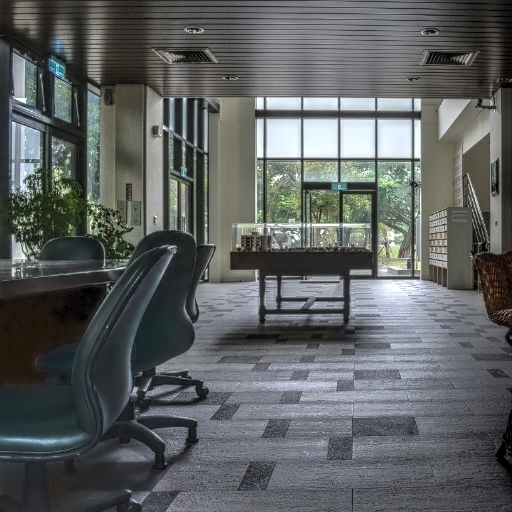}
    \includegraphics[width=0.15\linewidth, height=2.5cm]{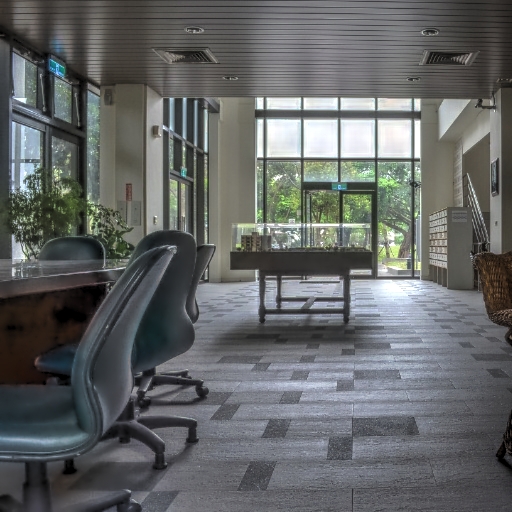}
    \includegraphics[width=0.15\linewidth, height=2.5cm]{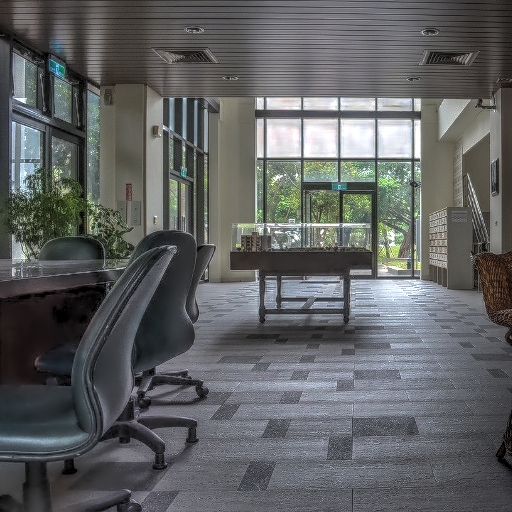}
    \includegraphics[width=0.15\linewidth, height=2.5cm]{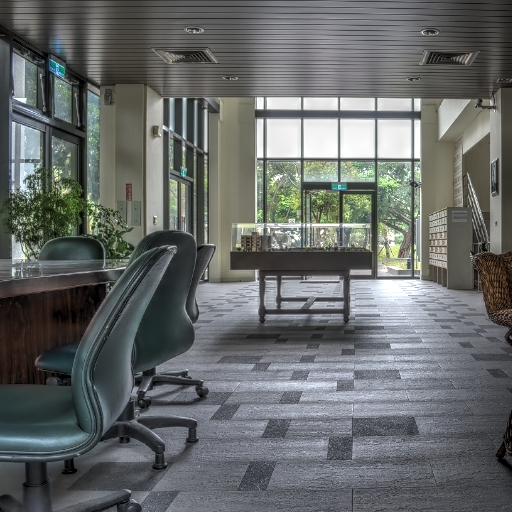}\\
    
    \vspace{0.5mm}
    
    \makebox[0pt][r]{\makebox[30pt]{\raisebox{30pt}{\rotatebox[origin=c]{0}{(e)}}}}%
    \includegraphics[width=0.15\linewidth, height=2.5cm]{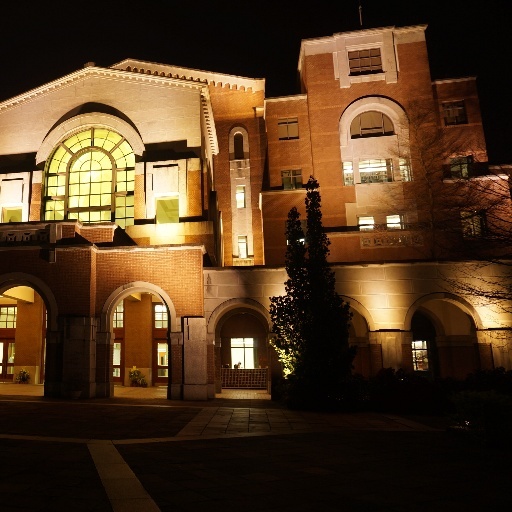}
    \includegraphics[width=0.15\linewidth, height=2.5cm]{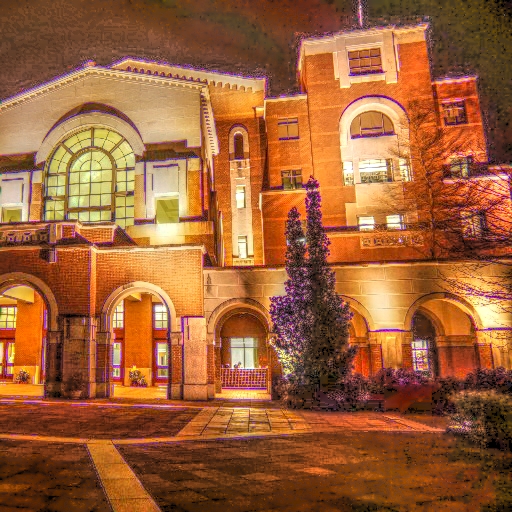}
    \includegraphics[width=0.15\linewidth, height=2.5cm]{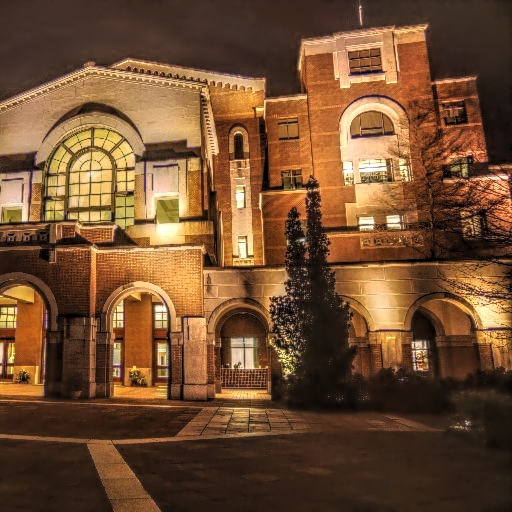}
    \includegraphics[width=0.15\linewidth, height=2.5cm]{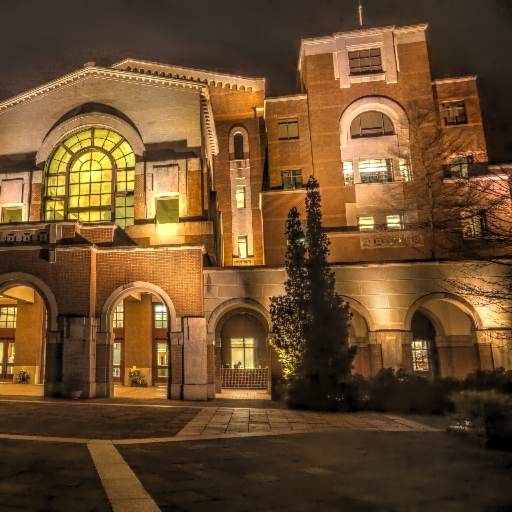}
    \includegraphics[width=0.15\linewidth, height=2.5cm]{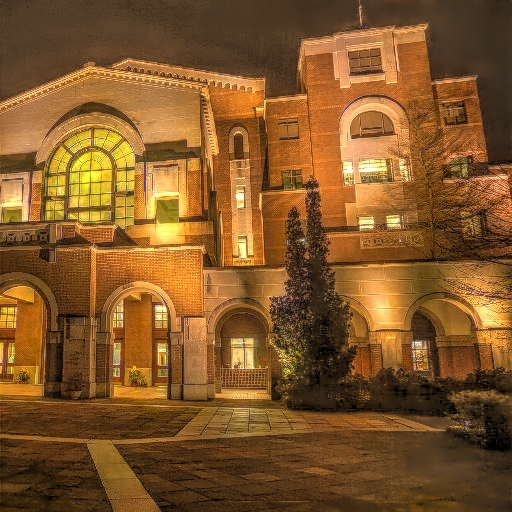}
    \includegraphics[width=0.15\linewidth, height=2.5cm]{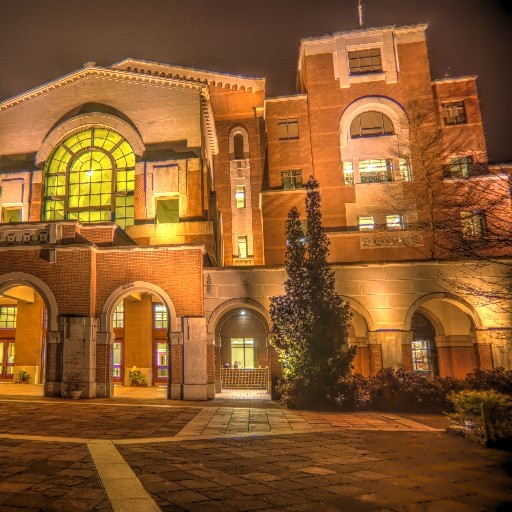}\\
    
    \vspace{0.5mm}
    
    \makebox[0pt][r]{\makebox[30pt]{\raisebox{30pt}{\rotatebox[origin=c]{0}{(f)}}}}%
    \includegraphics[width=0.15\linewidth, height=2.5cm]{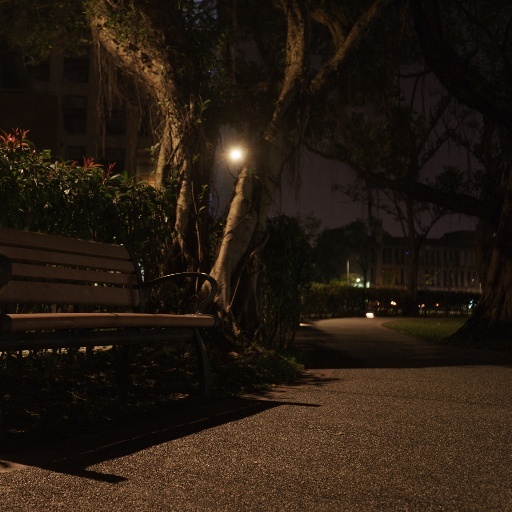}
    \includegraphics[width=0.15\linewidth, height=2.5cm]{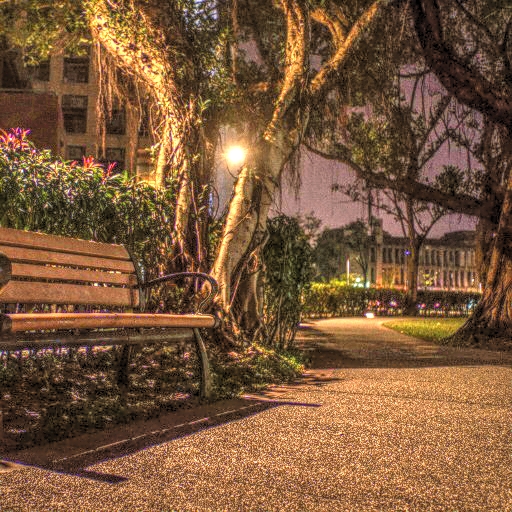}
    \includegraphics[width=0.15\linewidth, height=2.5cm]{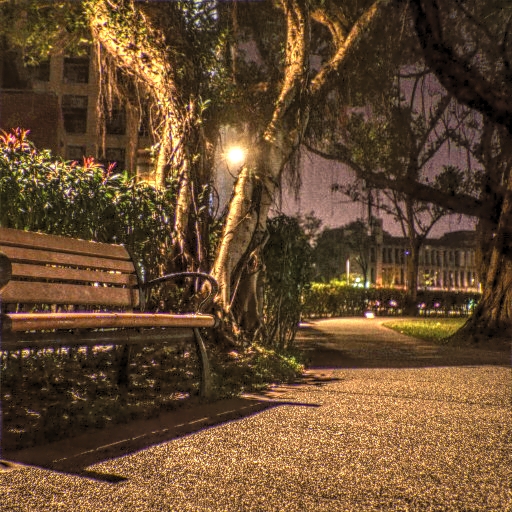}
    \includegraphics[width=0.15\linewidth, height=2.5cm]{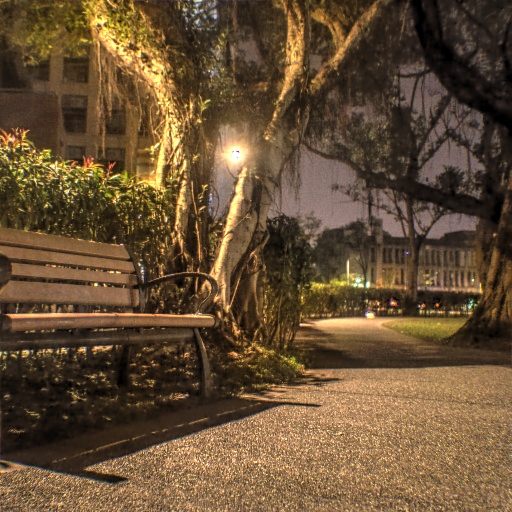}
    \includegraphics[width=0.15\linewidth, height=2.5cm]{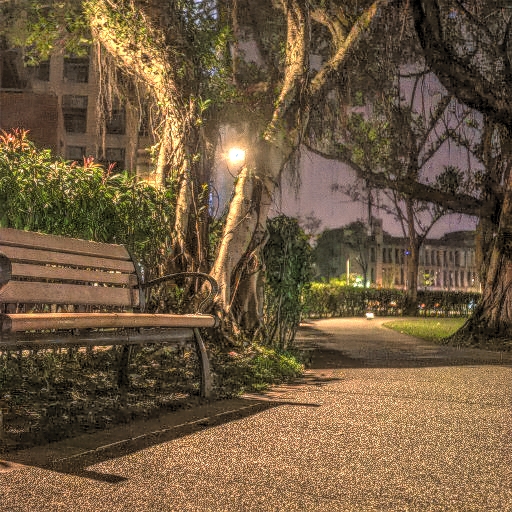}
    \includegraphics[width=0.15\linewidth, height=2.5cm]{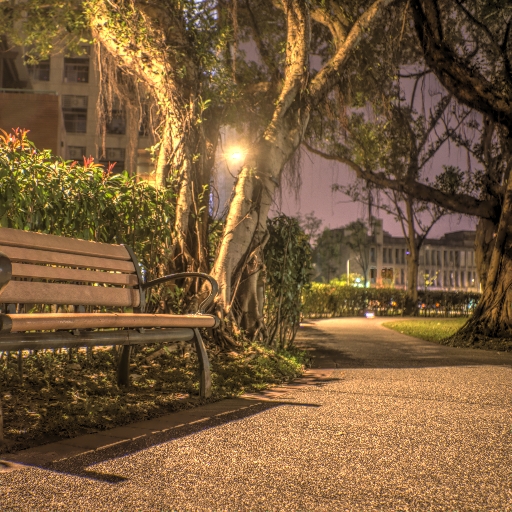}\\
    
    \vspace{0.5mm}
    
    \makebox[0pt][r]{\makebox[30pt]{\raisebox{30pt}{\rotatebox[origin=c]{0}{(g)}}}}%
    \includegraphics[width=0.15\linewidth, height=2.5cm]{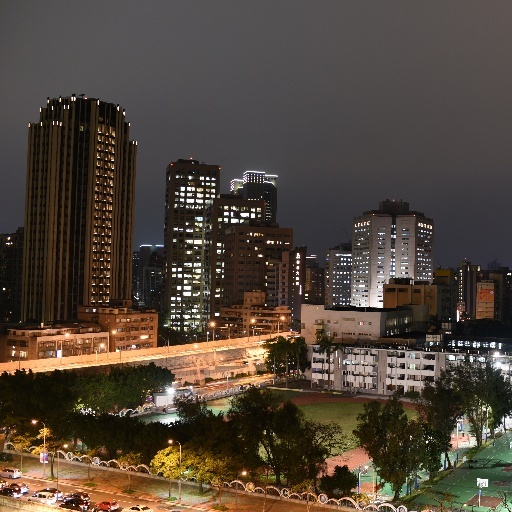}
    \includegraphics[width=0.15\linewidth, height=2.5cm]{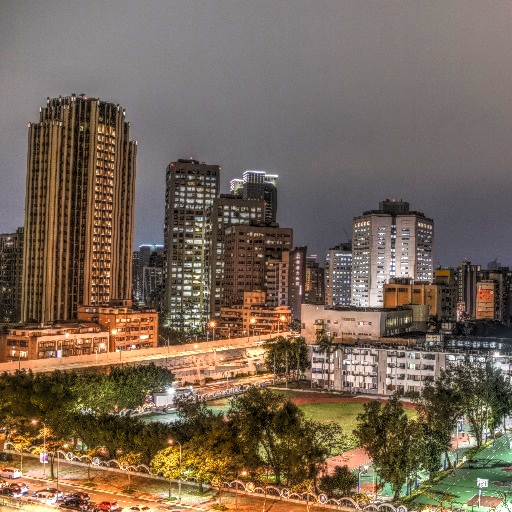}
    \includegraphics[width=0.15\linewidth, height=2.5cm]{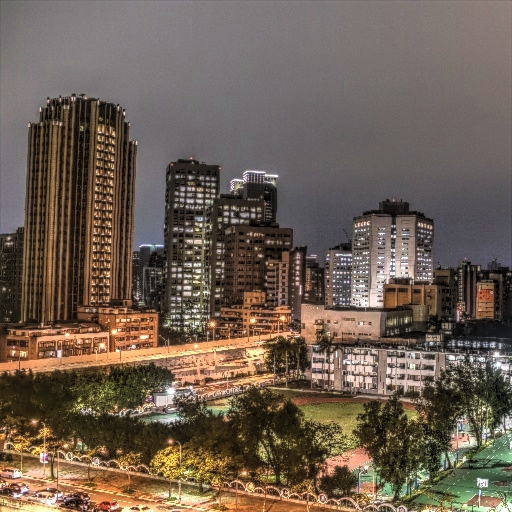}
    \includegraphics[width=0.15\linewidth, height=2.5cm]{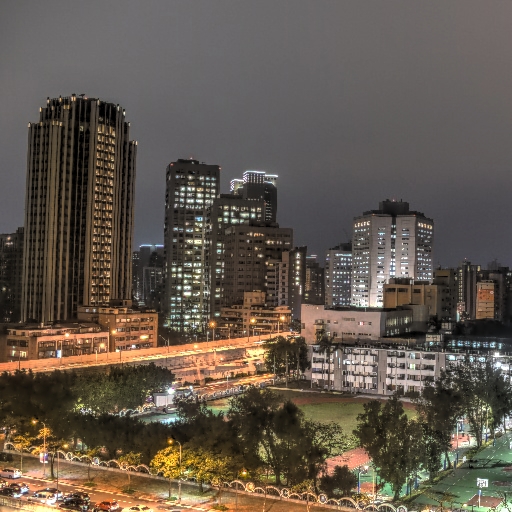}
    \includegraphics[width=0.15\linewidth, height=2.5cm]{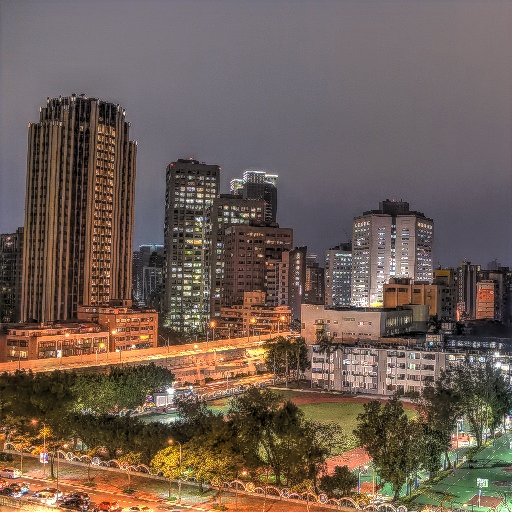}
    \includegraphics[width=0.15\linewidth, height=2.5cm]{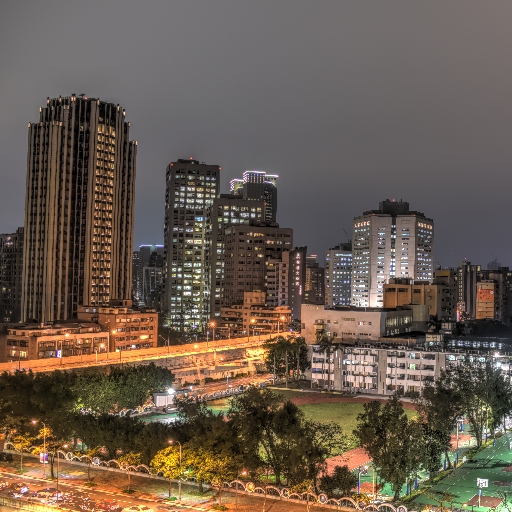}\\

    \vspace{1.5mm}

    \caption{Qualitative evaluation of our method against state-of-the-art methods. Rows (a) - (c) and (d) - (g) are samples from the HDR-EYE and HDR-REAL test sets respectively.}
    \label{fig:soft_segment}
\end{center}
\end{figure*}

\section{Experimental Results}
\label{experiments}

\textbf{Test Data.} We test the proposed model on the test set of  HDR-REAL \cite{liu2020singleimage} and HDR-EYE \cite{VISUALATTENTION} datasets. The test set of HDR-REAL dataset consists of $1838$  LDR-HDR image pairs. The HDR-Eye dataset consists of $46$ LDR-HDR image pairs. HDR images in these datasets are created by combining multi-exposure bracketed images captured with several cameras. All the input LDR images in our test set are in the JPEG format with each of size $512 \times 512$.

\textbf{Comparisons with state-of-the-art methods.} We compare our proposed method against some recent single-image based HDR reconstruction approaches, namely DrTMO \cite{10.1145/3130800.3130834}, HDRCNN \cite{eilertsen2017hdr}, ExpandNet \cite{marnerides2018expandnet}, and SingleHDR \cite{liu2020singleimage}. We directly perform the inference on their publicly available pre-trained models to obtain the output HDR images. These output images are quantitatively and qualitatively compared with the corresponding ground truth HDR images.

\subsection{Quantitative Comparison}

We perform the quantitative comparison of our proposed method over three metrics: PSNR, SSIM \cite{1284395}, and HDR-VDP $2.2$ \cite{article34}. The PSNR metric is commonly used to measure the reconstruction quality after lossy compression in an image. SSIM measures the similarity between two images based on their structural information. We apply the $\mu$ tonemapping operator to images before calculation of PSNR and SSIM metrics. Both PSNR and SSIM consider the low-level differences between the reconstructed output and the ground truth. However, these two metrics do not correlate well with human perception \cite{wan2020bringing}. Therefore, we also use the HDR-VDP $2.2$ metric, which can estimate the probability of distortions being visible to the human eye. HDR-VDP $2.2$ works specifically with the wide range of luminance availed by HDR images. We apply preprocessing steps as suggested in \cite{marnerides2018expandnet} to normalize both the ground-truth HDR and predicted HDR images.

Table \ref{hdreyerealtable} shows the quantitative evaluation of our approach on the HDR-Eye and HDR-Real datasets. On the HDR-Eye dataset, our method gives the best PSNR score and second-best SSIM and HDR-VDP $2.2$ scores. Our method is slightly behind ExpandNet \cite{marnerides2018expandnet} and SingleHDR \cite{liu2020singleimage} in the SSIM and HDR-VDP $2.2$ metrics respectively. On the HDR-Real dataset, our method ranks second-best in terms of all metrics. This time, SingleHDR \cite{liu2020singleimage} gives slightly better scores. The reason for their better performance may be the large synthetic training data ($103,010,400$ LDR images) created using 600 exposure times and 171 different CRFs. Our model is trained on a relatively small dataset comprising only $11,770$ images. Moreover, our model is end-to-end trainable, whereas \cite{liu2020singleimage} is trained in a stage-wise manner. In all, our method gives better results against conventional methods \cite{10.1145/3130800.3130834,eilertsen2017hdr,marnerides2018expandnet} and comparable results with the state-of-the-art \cite{liu2020singleimage} method.

\subsection{Qualitative Comparison} Figure \ref{fig:soft_segment} compares the visual results of our method against state-of-the-art methods. In normally-exposed regions of all the images, all methods produce decent results. However, in underexposed and overexposed regions, we always perform better than HDRCNN \cite{eilertsen2017hdr} and ExpandNet \cite{marnerides2018expandnet} and comparable to SingleHDR \cite{liu2020singleimage} method. As we see in Figure \ref{fig:soft_segment}\,(a) and \ref{fig:soft_segment}\,(b), our method generates more realistic textures in the building walls and ground respectively. This can be attributed to the adversarial and perceptual losses used in training. In Figure \ref{fig:soft_segment}\,(b) and \ref{fig:soft_segment}\,(c), overexposed regions have been very well hallucinated by our method. Our method produces comparable results to \cite{liu2020singleimage} in underexposed indoor scenes as observed in \ref{fig:soft_segment}\,(d). Our method also recovers more plausible lighting in the night scenes than \cite{liu2020singleimage} as observed in Figure \ref{fig:soft_segment}\,(e), \ref{fig:soft_segment}\,(f) and \ref{fig:soft_segment}\,(g). The sharpness and better color accuracy of our results are due to the conditional GAN framework. Note that we use the $vibrant$ mode of Photomatix \cite{phomat} software as a tonemapper to display the HDR images in qualitative comparison. This tonemapper is different from the $\mu$ tonemapper used in training. We recommend the readers to refer to supplementary material for the full-resolution view of the images illustrated in Figure \ref{fig:soft_segment}.

\subsection{Ablation Studies}

We validate the architecture of Attention Recurrent Residual U-Net used in our subnetworks by comparing with its following variant architectures:
\begin{itemize}
    \item \textbf{AttnR2\_UNet.} It is the full model of the architecture used in our subnetworks.
    \item \textbf{R2\_UNet.} In this model, we remove the attention module from the AttnR2\_UNet architecture.
    \item \textbf{Attn\_UNet.} In this model, we remove the recurrent residual module from the AttnR2\_UNet architecture and only the attention module is used.
\end{itemize}

\begin{figure}[h]
\;\;\;\;\; Input \;\;\;\;\;\;\;\;\;\;\;\;\;  R2\_UNet \;\;\;\;\;\;  Attn\_UNet \;\;\;\;\; AttnR2\_UNet\\
\vspace{-0.5mm}
\begin{center}
    
    \includegraphics[width=0.24\linewidth, height=2.0cm]{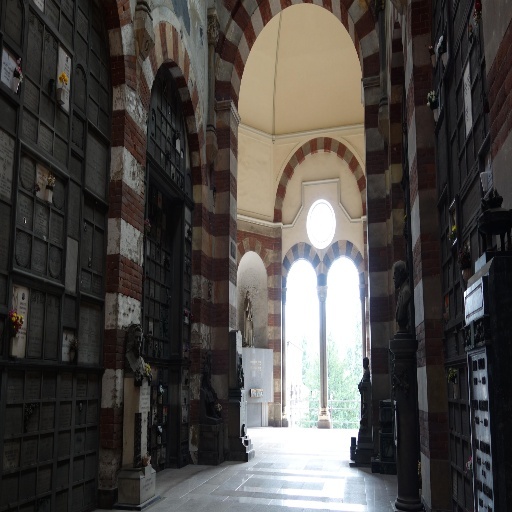}
    \includegraphics[width=0.24\linewidth,height=2.0cm]{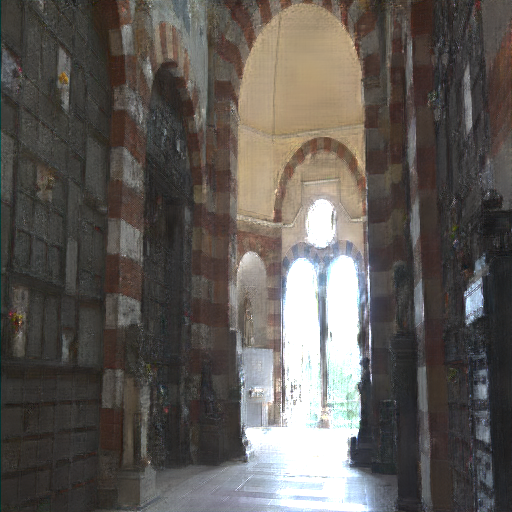}
    \includegraphics[width=0.24\linewidth, height=2.0cm]{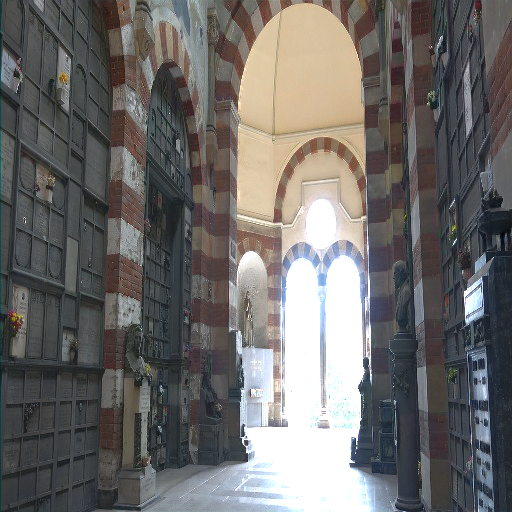}
    \includegraphics[width=0.24\linewidth, height=2.0cm]{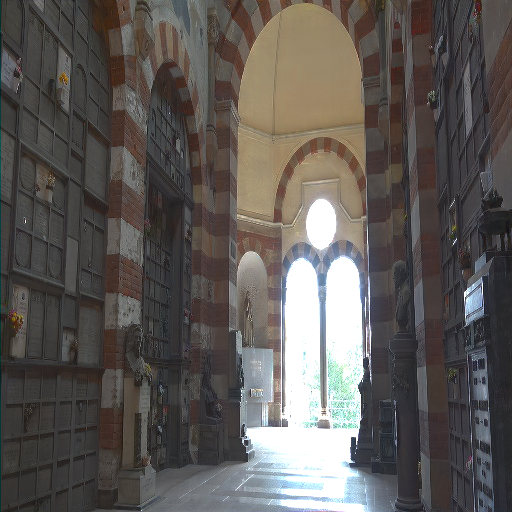}\\

    \vspace{1.5mm}

    \caption{Visual comparison of variants of AttnR2\_UNet.}
    \label{ablationfigure}
\end{center}
\end{figure}

Table \ref{ablationtable} shows that AttnR2\_UNet yields superior results over its variant architectures. Since the attention module alone can only recover the ground-truth exposure to a limited extent, we also incorporated recurrent residual blocks in our architecture. The combination of attention and the recurrent residual modules helps AttnR2\_UNet to hallucinate sharper details in the saturated regions with fewer artifacts as shown in Figure \ref{ablationfigure}. 

\begin{table}[t]
\begin{center}
\resizebox{\columnwidth}{!}{%
\begin{tabular}{|c||l|l|l|}
\hline
Method     & \multicolumn{1}{c|}{PSNR$(\uparrow)$} & \multicolumn{1}{c|}{SSIM$(\uparrow)$} & \multicolumn{1}{c|}{VDP $2.2$$(\uparrow)$} \\ \hline\hline
R2\_UNet   & $16.65\pm4.68$  & $0.75\pm0.14$   & $51.05\pm5.18$ \\ \hline
Attn\_UNet & $17.28\pm5.04$ & $0.78\pm0.14$   & $50.63\pm5.32$  \\ \hline
AttnR2\_UNet       & \color{red}{$17.57\pm4.68$}& \color{red}{$0.78\pm0.13$} & \color{red}{$51.94\pm5.56$} \\ \hline
\end{tabular}
}
\end{center}
\caption{Quantitative evaluation of our approach against different architectures in HDR-EYE dataset.}
\label{ablationtable}
\end{table}

\begin{figure}[h]
\;\; Input LDR \;\;\;\;\;\;\;\;\;\;  w/o Refinement \;\;\;\;\; with Refinement
\vspace{-0.5mm}
\begin{center}
    \includegraphics[width=0.30\linewidth, height=2.5cm]{pics/input/00017.jpg}
    \includegraphics[width=0.30\linewidth, height=2.5cm]{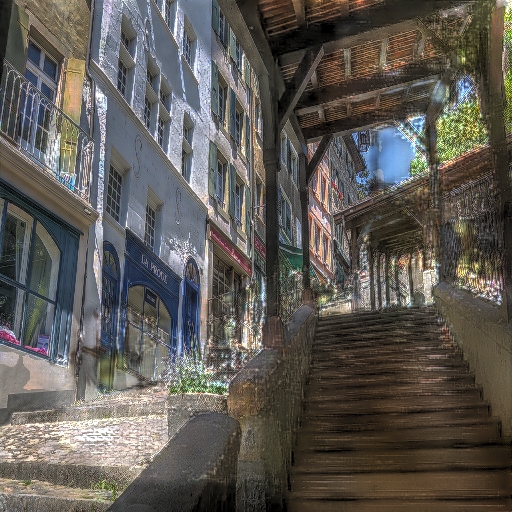}
    \includegraphics[width=0.30\linewidth, height=2.5cm]{pics/exp13/generated_hdr_00017.jpg}\\

    \vspace{1.5mm}

    \caption{Visual Comparison of results for with and without Refinement Net.}
    \label{refineablation}
\end{center}
\end{figure}

\textbf{Importance of Refinement Net.} In Figure \ref{refineablation}, we show the effectiveness of the Refinement Net in the overall pipeline. Refinement Net deals with the irregularities in the output of the OERC module. This helps the overall model to generate a high-quality HDR image without any artifacts. As shown in Figure \ref{refineablation}, our model without Refinement Net produces blurry artifacts in the output HDR image.


\subsection{Limitations}
\begin{figure}[h]
\;\; Input LDR \;\;\;\;\;\;\;\;\;\;\;\;\;\;\;\;\;  Ours \;\;\;\;\;\;\;\;\;\;\;\;\;\;  Ground truth
\vspace{-0.5mm}
\begin{center}
    
    \includegraphics[width=0.30\linewidth, height=2.5cm]{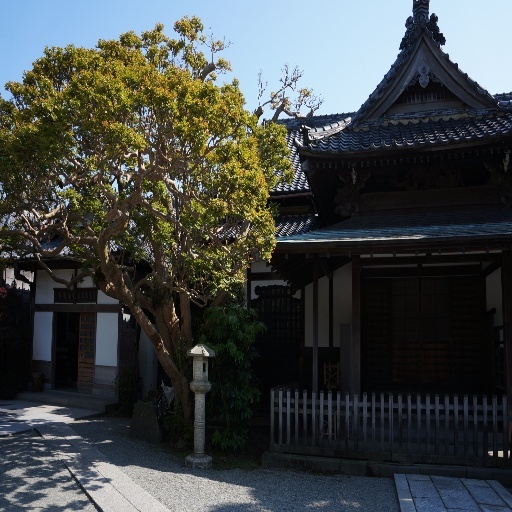}
    \includegraphics[width=0.30\linewidth,height=2.5cm]{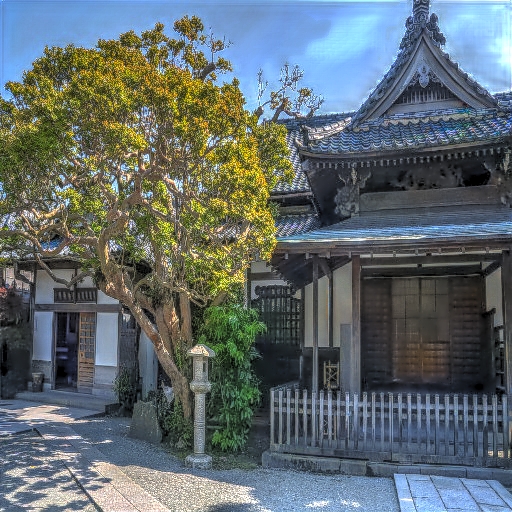}
    \vspace{0.5mm}
    \includegraphics[width=0.30\linewidth, height=2.5cm]{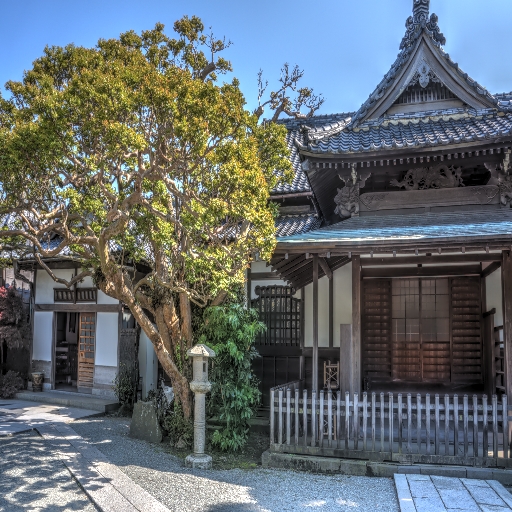}\\
    \includegraphics[width=0.30\linewidth, height=2.5cm]{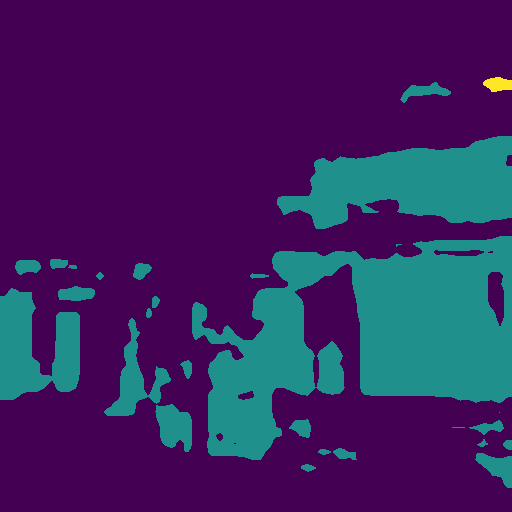}
    \includegraphics[width=0.30\linewidth, height=2.5cm]{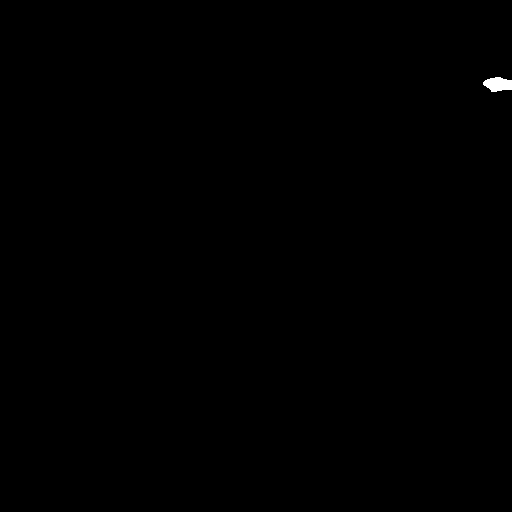}\\

    \vspace{1.5mm}

    \caption{Failure Case.\;\;In the bottom row, the segmentation mask is unable to predict the overexposed regions in the bright sky. Since the OE mask is not correctly estimated, our method sometimes produces artifacts in the overexposed regions.}
    \label{Limitations}
\end{center}
\end{figure}

Figure \ref{Limitations} shows a failure case of our method. Generally, our method properly hallucinates the saturated regions in an image. However, sometimes it fails to reconstruct the missing details and generate artifacts in those areas (\eg \, refer to the regions in the sky in Figure \ref{Limitations}). These artifacts occur when the pre-trained segmentation model does not correctly estimate the OE mask.

\section{Conclusion}
\label{conclusion}
We presented a conditional GAN based approach to reconstruct an HDR image from a single LDR image. Specifically, we formulated the LDR-HDR pipeline as a sequence of three steps, linearization, over-exposure correction, and refinement. We tackled each of these problems using separate networks and trained them jointly with a combination of objective functions. Linearization module transfers the input LDR image into the HDR domain and also recovers detail in underexposed regions. In order to explicitly drive the training towards generating details in the saturated areas of overexposed regions, we also incorporated an Over Exposure mask estimated by a pre-trained segmentation model. Refinement module helps to generate an artifact-free HDR image. Finally, we showed through experiments and analysis that our cGAN based framework is effective in learning a complex mapping like LDR to HDR translation.

\begin{acks}
This research was supported by the SERB Core Research
Grant.
\end{acks}

\bibliographystyle{ACM-Reference-Format}
\bibliography{ICVGIP21-CameraReady-Template}





\end{document}